\documentclass[sigconf,nonacm]{acmart}

\usepackage{multirow}
\usepackage{multicol}
\usepackage{subcaption}
\usepackage{cleveref}
\usepackage{balance}
\usepackage{microtype}
\usepackage{adjustbox}
\usepackage{algpseudocode}
\usepackage{colortbl}
\usepackage{enumitem}
\usepackage[most]{tcolorbox}
\usepackage{algorithm}

\algrenewcommand\algorithmicrequire{\textbf{Input:}}
\algrenewcommand\algorithmicensure{\textbf{Output:}}

\AtBeginDocument{%
  }

\setcopyright{none}
\renewcommand\footnotetextcopyrightpermission[1]{}

\title{One Pretrained Decision Model Is Enough for Training-Free Generalist Graph Anomaly Detection}

\author{Xinyu Zhao}
\affiliation{%
  \institution{School of Computer Science and Engineering, Beihang University}
  \city{Beijing}
  \country{China}
  }
\email{xyzhao@buaa.edu.cn}

\author{Qingyun Sun}
\authornote{Corresponding Author.}
\affiliation{%
  \institution{School of Computer Science and Engineering, Beihang University}
  \city{Beijing}
  \country{China}
  }
\email{sunqy@buaa.edu.cn}

\author{Ziwei Zhang}
\affiliation{%
  \institution{School of Computer Science and Engineering, Beihang University}
  \city{Beijing}
  \country{China}
  }
\email{zwzhang@buaa.edu.cn}

\author{Tianchen Zhu}
\affiliation{%
  \institution{School of Reliability and Systems Engineering, Beihang University}
  \city{Beijing}
  \country{China}
  }
\email{catezi@buaa.edu.cn}

\author{Daqing Li}
\affiliation{%
  \institution{School of Reliability and Systems Engineering, Beihang University}
  \city{Beijing}
  \country{China}
  }
\email{daqingl@buaa.edu.cn}

\author{Jianxin Li}
\affiliation{%
  \institution{School of Computer Science and Engineering, Beihang University}
  \city{Beijing}
  \country{China}
  }
\email{lijx@buaa.edu.cn}

\renewcommand{\shortauthors}{Xinyu Zhao et al.}

\begin{document}

\begin{abstract}
Graph anomaly detection (GAD) aims to identify irregular entities in graph-structured data, yet heterogeneous anomaly patterns and distributional differences across graphs challenge generalizable detection. Existing training-free approaches avoid task-specific optimization but primarily rely on predefined numerical scoring rules, leaving the semantic interpretation of heterogeneous anomaly evidence largely unexplored. In this paper, we propose Jev-GAD, a training-free generalist graph anomaly detection framework that bridges heterogeneous numerical evidence and pretrained decision mechanisms. To address heterogeneous anomaly patterns, we construct complementary structural, attribute, and neighborhood evidence and transform heterogeneous statistics into graph-relative anomaly percentiles. To move beyond predefined numerical scoring rules, we leverage a single frozen pretrained language model for evidence-conditioned anomaly assessment. To bridge the semantic gap between numerical evidence and pretrained models, we design a semantic Graph-to-State (G2S) transformation that organizes graph context and semantically labeled node evidence into structured decision states for non-generative option-logit readout. The resulting decision scores are integrated with four complementary statistical scores through equal-weight rank-based fusion, without GAD-specific gradient-based optimization. Extensive experiments on seven graph anomaly detection benchmarks spanning citation, social, and co-review networks demonstrate the effectiveness of Jev-GAD, achieving average AUROC and AUPRC scores of 89.18\% and 52.40\%, respectively. Further analyses validate the contributions of pretrained decision readout and semantic evidence grounding.
\end{abstract}



\maketitle

\section{Introduction}
\label{sec:introduction}

Graph Anomaly Detection (GAD) plays a crucial role in maintaining the trustworthiness of Web platforms by identifying suspicious entities and irregular interactions in online social networks, e-commerce systems, and recommendation services \cite{rayana2015collective,kumar2018rev2,zhang2022efraudcom}. In these environments, anomalous behaviors may manifest as unusual structural positions, atypical node attributes, or inconsistencies with neighboring entities \cite{ma2021comprehensive,qiao2025deep}. For instance, fraudulent users may disguise their attributes or relationships to evade detection \cite{dou2020enhancing,zheng2024friend}, while abnormal relational patterns can exhibit varying degrees of neighborhood heterophily \cite{gao2023addressing}. Such heterogeneous anomaly characteristics, together with substantial distributional differences across Web graphs, make it challenging to develop detectors that generalize across domains. Traditional GAD methods typically rely on reconstruction or contrastive learning objectives \cite{ding2019deep,fan2020anomalydae,liu2021anomaly}, while recent generalist approaches explore transferring anomaly detection capabilities to unseen graphs \cite{liu2024arc,niu2024zero,qiao2025anomalygfm,zhang2026ia,zhao2026zero}. Despite their effectiveness, many existing methods still require source-domain training or specialized adaptation. Training-free approaches \cite{zhou2025training,zhao2025freegad} offer an appealing alternative by exploiting intrinsic graph characteristics without GAD-specific parameter training, particularly when target-domain anomaly labels are unavailable and graph distributions vary across domains.

\begin{figure}[t]
    \centering
    \includegraphics[width=\columnwidth]{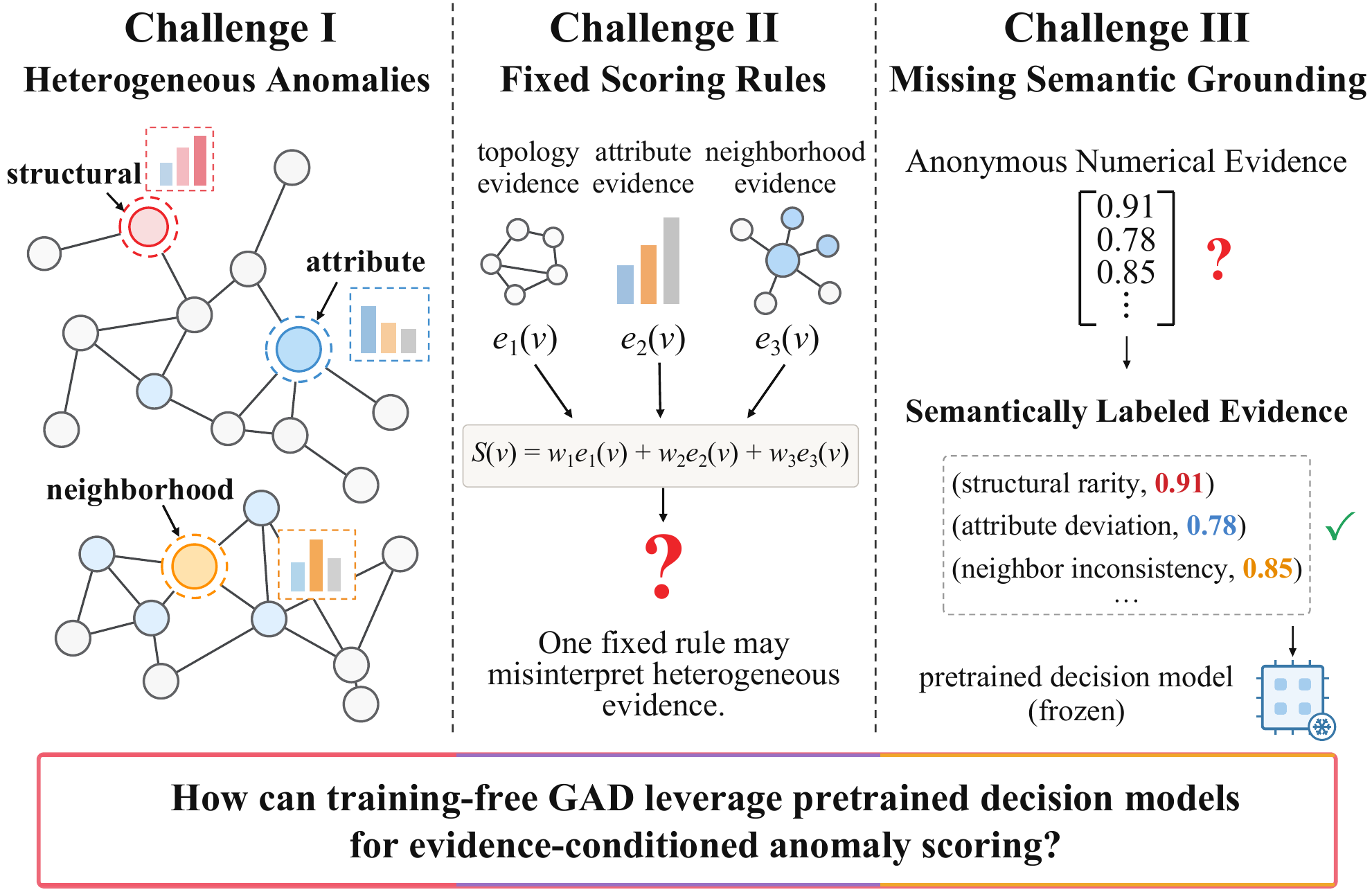}
    \caption{Motivation of Jev-GAD: heterogeneous anomalies, fixed scoring rules, and missing semantic grounding.}
    \label{fig:motivation}
\end{figure}

However, eliminating task-specific training does not necessarily eliminate the limitations of predefined anomaly scoring mechanisms. As illustrated in Fig.~\ref{fig:motivation}, heterogeneous graph anomalies give rise to distinct forms of structural, attribute, and neighborhood evidence, each reflecting different characteristics of anomalous behavior. Existing training-free methods commonly transform such evidence into anomaly scores through predefined numerical criteria \cite{zhou2025training,zhao2025freegad}. Although these criteria can effectively capture particular irregularities, they do not explicitly condition their assessments on the semantic meanings of different evidence types. For instance, structural rarity and neighborhood inconsistency may both indicate anomalous behavior, yet their implications can vary with the underlying graph context. Consequently, a fixed numerical scoring mechanism may struggle to accommodate heterogeneous, context-dependent anomaly patterns. This limitation raises a central question: \emph{how can training-free GAD leverage pretrained decision models for evidence-conditioned anomaly scoring?}

Meanwhile, pretrained language models have shown promising capabilities in evaluation, classification, and structured decision-making beyond conventional text generation \cite{zheng2023judging,kim2024prometheus,deusser2026evaluating,ren2026open}. These advances suggest the potential of leveraging a single frozen pretrained model for graph anomaly assessment. However, directly applying such models introduces a fundamental \emph{semantic gap}: graph anomaly evidence primarily consists of numerical statistics whose interpretations depend on structural, attribute, and relational contexts. Even after graph-relative normalization, anonymous numerical values cannot distinguish structural rarity from attribute deviation or neighborhood inconsistency, making it difficult for pretrained models to interpret the anomaly characteristics represented by different evidence dimensions. This observation motivates our key insight: \emph{by associating graph-relative numerical evidence with explicit anomaly semantics, frozen pretrained models can perform evidence-conditioned anomaly assessment without graph-specific parameter learning.} Such semantic grounding provides a bridge between heterogeneous numerical graph evidence and pretrained decision mechanisms.

Building upon this insight, we propose \textbf{Jev-GAD}, a training-free generalist graph anomaly detection framework that bridges heterogeneous graph evidence and pretrained decision mechanisms. To accommodate diverse anomaly patterns and graph distributions, Jev-GAD constructs complementary structural, attribute, and neighborhood evidence and transforms heterogeneous statistics into graph-relative anomaly percentiles.
To move beyond predefined numerical scoring rules, we leverage a single frozen pretrained language model for evidence-conditioned anomaly assessment. To bridge the semantic gap between numerical evidence and pretrained models, we introduce a semantic \emph{Graph-to-State} (G2S) transformation that deterministically organizes graph context and semantically labeled node evidence into structured decision states. The frozen model subsequently evaluates anomalous and normal alternatives through non-generative option-logit readout, producing continuous anomaly scores without task-specific parameter updates. In parallel, four complementary statistical scorers assess numerical irregularities, and their scores are integrated with pretrained decision scores through equal-weight rank-based fusion. Extensive experiments on seven benchmarks spanning citation, social, and co-review networks demonstrate the effectiveness of Jev-GAD, achieving average AUROC and AUPRC scores of 89.18\% and 52.40\%, respectively, surpassing the strongest baseline macro-averages by 7.09 and 15.44 percentage points. Further analyses validate the contributions of pretrained decision readout, semantic evidence grounding, and complementary statistical scoring.
Our contributions are summarized as follows:

\begin{itemize}
    \item We propose \textbf{Jev-GAD}, a training-free generalist graph anomaly detection framework that leverages a single frozen pretrained model for evidence-conditioned anomaly scoring beyond predefined numerical rules.

    \item We introduce \textbf{Graph-to-State (G2S)}, which transforms heterogeneous numerical evidence into semantically grounded decision states, enabling non-generative pretrained decision readout and its integration with complementary statistical scoring.

    \item Extensive experiments on seven graph anomaly detection benchmarks demonstrate the effectiveness of Jev-GAD, while comprehensive analyses validate the contributions of pretrained decisions, semantic grounding, and complementary statistical evidence.
\end{itemize}

\section{Related Work}

\subsection{Graph Anomaly Detection}
Traditional GAD methods identify anomalous nodes through residual modeling, reconstruction, or contrastive learning, as exemplified by Radar, DOMINANT, AnomalyDAE, CoLA, ANEMONE, CONAD, and GAD-NR~\cite{li2017radar,ding2019deep,fan2020anomalydae,liu2021anomaly,jin2021anemone,xu2022contrastive,roy2024gad}. Cluster-aware methods further exploit community-level patterns for graph anomaly detection~\cite{zheng2025cluster}. Together, these approaches capture different levels of anomaly evidence, ranging from node attributes and neighborhood relations to higher-order community structure. Large-scale benchmarking efforts further reveal the diverse characteristics and performance of unsupervised graph anomaly detectors across attributed networks~\cite{liu2022bond}. Such diversity motivates combining complementary structural, attribute, and neighborhood evidence. Recent generalist methods, including ARC, UNPrompt, AnomalyGFM, IA-GGAD, and GAD-MoRE, further explore transferable anomaly detection across diverse graph domains \cite{liu2024arc,niu2024zero,qiao2025anomalygfm,zhang2026ia,zhao2026zero}. However, many of these approaches still require task-specific training or adaptation. To reduce this dependency, training-free methods such as TFGAD and FreeGAD leverage intrinsic graph statistics and predefined scoring mechanisms to identify anomalies \cite{zhou2025training,zhao2025freegad}. 
Despite their effectiveness, these methods primarily rely on statistical scoring criteria, which may struggle to capture context-dependent anomaly patterns, particularly when different evidence types interact.

\subsection{Pretrained Decision Models}
Recent advances in large language models (LLMs) have enabled automated evaluation and decision-making beyond text generation \cite{zheng2023judging,kim2024prometheus}. Unlike free-form generative judging, non-generative decision readout directly obtains option-level scores conditioned on predefined questions and contextual states. Recent studies have systematically evaluated Jev's zero-shot decision capabilities across diverse classification and reasoning tasks \cite{deusser2026evaluating,rafe2026benchmarking}. Further investigations demonstrate its potential for efficient preference evaluation, scientific decision-making, and semantic query processing \cite{li2026jev,deng2026jev,wang2026prune}. Meanwhile, Open-Jev explores single-pass decision readout through option logits, while recent research examines the stability of such non-generative prediction mechanisms \cite{ren2026open,li2026readout}. Option-level scores also allow pretrained decisions to be integrated directly with numerical anomaly scorers. Despite these advances, applying pretrained decision mechanisms to heterogeneous graph evidence remains relatively underexplored.

\section{Problem Formulation}

We consider node-level graph anomaly detection on an attributed graph $\mathcal{G}=(\mathcal{V},\mathcal{E},\mathbf{X}_0)$, where $\mathcal{V}$ and $\mathcal{E}$ denote the node and edge sets, respectively, and $\mathbf{X}_0\in\mathbb{R}^{N\times D_0}$ is the node feature matrix. Let $\mathbf{A}\in\{0,1\}^{N\times N}$ denote the adjacency matrix, where $N=|\mathcal{V}|$. The objective is to construct an anomaly scoring function $f:\mathcal{V}\rightarrow\mathbb{R}$ that assigns higher scores to anomalous nodes. We focus on the training-free generalist setting, where the detector is directly applied to unseen target graphs $\mathcal{D}_{\mathrm{test}}=\{\mathcal{G}^{(t)}\}_{t=1}^{T}$ without access to their anomaly labels $\mathbf{y}^{(t)}\in\{0,1\}^{N_t}$ or GAD-specific gradient-based training. The pretrained decision model $g_{\theta}$ remains frozen, and $\mathbf{y}^{(t)}$ is used solely for evaluation.
\begin{figure*}[t]
    \centering
    \includegraphics[width=0.96\textwidth]{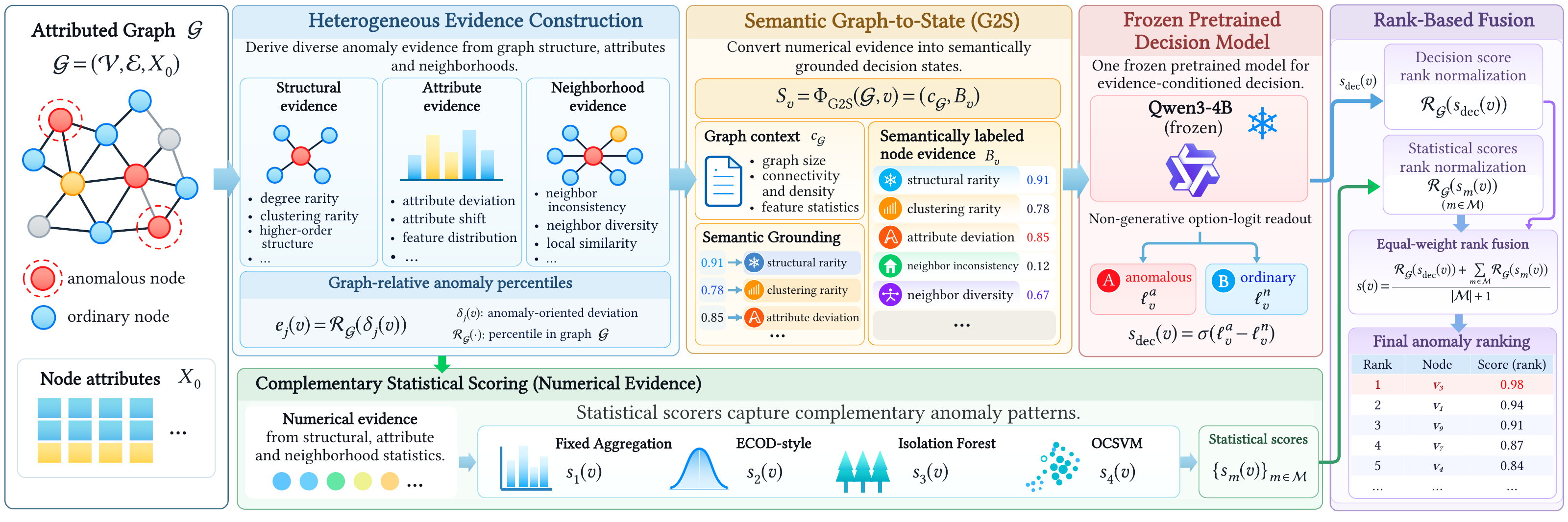}
    \caption{Overview of Jev-GAD. Heterogeneous anomaly evidence is organized into semantically grounded decision states for a single frozen pretrained model. Complementary statistical scoring and rank-based fusion produce the final anomaly ranking without GAD-specific gradient-based optimization.}
    \label{fig:framework}
\end{figure*}

\section{Methodology}
\label{sec:method}

As illustrated in Figure~\ref{fig:framework}, Jev-GAD constructs graph-relative anomaly evidence from structure, attributes, and neighborhoods, organizes selected evidence into semantically grounded states for pretrained decision inference, and integrates the resulting decision scores with complementary statistical assessments through rank-based fusion.

\subsection{Heterogeneous Graph Evidence Construction}
\label{sec:evidence}

Graph anomalies manifest as structural irregularities, attribute deviations, and neighborhood inconsistencies. We construct complementary graph evidence to characterize these heterogeneous anomaly patterns.

\noindent\textbf{Complementary Graph Evidence.} Given an attributed graph $\mathcal{G}=(\mathcal{V},\mathcal{E},\mathbf{X}_0)$, we consider three categories of anomaly evidence.

Structural evidence characterizes how a node differs from the typical topological organization of the graph, including its connectivity, local clustering, core membership, and centrality. Attribute evidence measures deviations from the global feature distribution, while neighborhood evidence captures inconsistencies between a node and its local relational context.

To complement the original attributes, we additionally construct an aligned feature view using the training-free multi-curvature alignment procedure of~\cite{zhao2026zero}. Attribute and neighborhood evidence are extracted from both the original and aligned views, allowing different representations to expose potentially distinct anomaly characteristics. The detailed evidence configuration is provided in Appendix~\ref{app:evidence}.

\noindent\textbf{Graph-Relative Evidence.} A central difficulty in combining heterogeneous evidence is that different statistics have incompatible scales and interpretations. For instance, a large feature distance may indicate an anomaly, while structural irregularities can appear at either extreme of a statistic's distribution. Moreover, the same numerical value may have different implications across graphs with different structural and attribute distributions.

We address this issue by converting each statistic into a graph-relative measure of atypicality. Let $q_j(v)$ denote the $j$-th statistic of node $v$, and let $\delta_j(v)$ be its anomaly-oriented deviation. For statistics with potentially anomalous upper and lower tails, $\delta_j(v)$ measures the deviation from the graph-wide median; for distance-based statistics, larger deviations are retained directly. The normalized evidence is defined as
\begin{equation}
    e_j(v)=\mathcal{R}_G\bigl(\delta_j(v)\bigr),
    \label{eq:evidence_normalization}
\end{equation}
where $\mathcal{R}_G$ computes the empirical percentile rank among all nodes in the target graph.

This transformation provides a common numerical interpretation: evidence values lie in $[0,1]$, with larger values indicating greater atypicality relative to the graph population. The resulting 16-dimensional numerical evidence representation is
\begin{equation}
    \mathbf{e}_v=\left[\mathbf{e}_v^{\mathrm{str}};\mathbf{e}_v^{\mathrm{raw}};\mathbf{e}_v^{\mathrm{align}}\right],
    \label{eq:graph_evidence}
\end{equation}
where the components comprise four structural indicators and six attribute and neighborhood indicators from each feature view. This representation captures complementary anomaly patterns and provides the numerical basis for statistical scoring.

\subsection{Semantic Graph-to-State Representation}
\label{sec:graph_to_state}

Although graph-relative normalization makes heterogeneous evidence numerically comparable, numerical comparability alone does not make the evidence interpretable to a pretrained model. For example, a high percentile may describe either an unusual structural position or a strong disagreement with neighboring nodes. These observations can have different implications for anomaly assessment, yet their meanings cannot be recovered from anonymous numerical values alone.

This motivates a semantic interface that preserves the correspondence between each numerical observation and the graph characteristic it describes. We introduce a semantic \emph{Graph-to-State} (G2S) transformation that organizes graph evidence into structured decision states with explicit semantic correspondence.

\noindent\textbf{Semantic Evidence Grounding.} Let $\mathcal{I}=(j_1,\ldots,j_7)$ denote the ordered indices of the seven indicators selected for decision readout. For each selected indicator $j_k$, we associate its normalized value $e_{j_k}(v)$ with a semantic descriptor $\psi_{j_k}$ specifying the corresponding anomaly characteristic.

The resulting evidence state is represented as
\begin{equation}
    \mathcal{B}_v=\left[(\psi_{j_k},e_{j_k}(v))\right]_{k=1}^{7}.
    \label{eq:semantic_grounding}
\end{equation}

Unlike an anonymous feature vector, $\mathcal{B}_v$ retains both the magnitude and interpretation of each observation. Its seven entries follow a fixed order across neighborhood, structural, and attribute groups, allowing the decision model to distinguish different sources of irregularity.

For example, a node may exhibit ordinary structural characteristics but strong neighborhood inconsistency. Preserving these distinctions enables the pretrained model to evaluate the supplied evidence with explicit knowledge of what each numerical value represents, rather than relying on feature positions alone.

\noindent\textbf{Graph-to-State Construction.} Node-level evidence is further accompanied by auxiliary graph-level context $\mathbf{c}_G$, which summarizes properties of the observed graph, including its size, connectivity, and feature characteristics. We define the resulting state as
\begin{equation}
    \mathcal{S}_v=\Phi_{\mathrm{G2S}}(G,v)=\left(\mathbf{c}_G,\mathcal{B}_v\right),
    \label{eq:graph_to_state}
\end{equation}
where $\Phi_{\mathrm{G2S}}$ denotes the Graph-to-State transformation.

The state is serialized into a structured input with a fixed structure across nodes and graphs. Its node-level entries are grouped by anomaly characteristic and retain their descriptive semantics. The graph-level context provides auxiliary information about the observed graph.

Two properties are particularly important. First, semantic descriptors and numerical evidence remain aligned across graphs. Second, graph-relative normalization ensures that the values share a consistent direction and represent relative atypicality rather than graph-specific raw magnitudes.

These properties enable a frozen pretrained model to access heterogeneous graph evidence without graph-specific representation learning. Semantic correspondence is examined in Sec.~\ref{sec:semantic_analysis}.

\subsection{Pretrained Decision Readout}
\label{sec:decision_readout}

Given the semantically grounded state $\mathcal{S}_v$, Jev-GAD employs a frozen pretrained model to make evidence-conditioned anomaly decisions through non-generative option-logit readout.

\noindent\textbf{Evidence-Conditioned Decision.} We formulate anomaly assessment as a binary choice between anomalous and normal behavior. Each decision query combines a fixed decision criterion, two candidate options, and the node-specific state $\mathcal{S}_v$.

The criterion specifies how the evidence should be interpreted: larger normalized values represent stronger graph-relative deviations, while the accompanying semantic descriptions identify the nature of those deviations. The decision criterion and candidate options remain unchanged across nodes and graphs, whereas the evidence state varies with the node being evaluated.

This formulation allows the pretrained model to receive multiple anomaly characteristics as explicit decision evidence, rather than requiring an additional graph-specific classifier to be trained from labeled examples.

\noindent\textbf{Direct Option Readout.} Instead of generating a textual judgment, we adopt a non-generative decision readout inspired by single-pass decision mechanisms~\cite{ren2026open}. Specifically, a frozen pretrained language model processes the decision query, and we extract the output logits corresponding to the two fixed, single-token options.

Let $\ell_v^{a}$ and $\ell_v^{n}$ denote the logits associated with anomalous and normal decisions, respectively. The decision score is
\begin{equation}
    s_{\mathrm{dec}}(v)=\sigma\left(\ell_v^{a}-\ell_v^{n}\right),
    \label{eq:decision_score}
\end{equation}
where $\sigma(\cdot)$ is the sigmoid function.

A larger score indicates a stronger preference for the anomalous option and provides a continuous signal for anomaly ranking.

A single forward pass produces the decision score without autoregressive generation or parameter updates.

\begin{table*}[t]
\centering
\caption{AUROC comparison on seven graph anomaly detection benchmarks (\%). Results are reported as mean $\pm$ standard deviation. Best and second-best mean values are highlighted in \textbf{bold} and \underline{underline}, respectively.}
\label{tab:overall_auroc}

\begingroup
\setlength{\tabcolsep}{1.5pt}
\renewcommand{\arraystretch}{1.13}
\normalsize

\newcommand{\ms}[2]{#1{\footnotesize$\pm$#2}}
\newcommand{\venue}[1]{{\footnotesize(\textit{#1})}}

\begin{tabular*}{\textwidth}{@{\extracolsep{\fill}}lcccccccc@{}}
\toprule
\textbf{Method (Venue)}
& \textbf{ACM}
& \textbf{Amazon}
& \textbf{BlogCatalog}
& \textbf{CiteSeer}
& \textbf{Cora}
& \textbf{Facebook}
& \textbf{Weibo}
& \textbf{Avg.} \\
\midrule

\multicolumn{9}{@{}l}{\emph{Supervised Methods}} \\

GCN \venue{ICLR'17}
& \ms{47.69}{1.48}
& \ms{41.10}{4.24}
& \ms{43.83}{2.39}
& \ms{53.83}{4.01}
& \ms{51.99}{3.20}
& \ms{46.09}{18.88}
& \ms{37.64}{13.84}
& 46.02 \\

GAT \venue{ICLR'18}
& \ms{46.54}{4.71}
& \ms{55.35}{5.69}
& \ms{49.32}{1.70}
& \ms{44.58}{8.25}
& \ms{44.65}{6.20}
& \ms{53.96}{26.14}
& \ms{35.50}{6.91}
& 47.13 \\

BWGNN \venue{ICML'22}
& \ms{67.43}{2.48}
& \ms{50.01}{0.04}
& \ms{69.91}{4.66}
& \ms{42.58}{2.86}
& \ms{39.75}{3.16}
& \ms{64.84}{4.17}
& \ms{50.08}{0.03}
& 54.94 \\

GHRN \venue{WWW'23}
& \ms{52.72}{1.01}
& \ms{63.52}{1.92}
& \ms{62.28}{2.42}
& \ms{54.11}{0.88}
& \ms{47.65}{2.03}
& \ms{54.04}{3.81}
& \ms{17.79}{5.00}
& 50.30 \\

\midrule
\multicolumn{9}{@{}l}{\emph{Unsupervised Methods}} \\

AnomalyDAE \venue{ICASSP'20}
& \ms{77.60}{1.03}
& \ms{58.38}{0.25}
& \ms{57.53}{0.30}
& \ms{82.00}{2.78}
& \ms{84.01}{1.19}
& \ms{10.87}{0.70}
& \ms{19.77}{1.14}
& 55.74 \\

CoLA \venue{TNNLS'21}
& \ms{49.30}{1.00}
& \ms{47.57}{3.02}
& \ms{49.50}{1.27}
& \ms{52.56}{1.36}
& \ms{49.35}{1.52}
& \ms{50.70}{5.49}
& \ms{45.25}{2.39}
& 49.17 \\

TAM \venue{NeurIPS'23}
& \ms{50.39}{0.08}
& \ms{52.37}{0.12}
& \ms{33.03}{0.28}
& \ms{46.44}{0.15}
& \ms{48.78}{0.17}
& \ms{54.56}{0.36}
& \ms{21.50}{0.30}
& 43.87 \\

\midrule
\multicolumn{9}{@{}l}{\emph{Generalist Methods}} \\

ARC (10-shot) \venue{NeurIPS'24}
& \ms{78.33}{1.13}
& \ms{57.95}{6.13}
& \ms{72.59}{1.10}
& \ms{82.31}{2.89}
& \ms{82.22}{2.34}
& \ms{64.94}{3.72}
& \ms{88.65}{0.82}
& 75.28 \\

AnomalyGFM \venue{KDD'25}
& \ms{38.91}{1.21}
& \ms{61.24}{5.21}
& \ms{40.93}{1.22}
& \ms{40.37}{1.37}
& \ms{40.93}{1.09}
& \ms{74.45}{2.36}
& \ms{44.43}{4.89}
& 48.75 \\

UNPrompt \venue{IJCAI'25}
& \ms{73.20}{0.36}
& \ms{72.14}{2.94}
& \ms{70.85}{0.50}
& \ms{59.53}{0.53}
& \ms{63.18}{1.29}
& \ms{60.64}{1.94}
& \ms{66.29}{1.38}
& 66.55 \\

IA-GGAD \venue{NeurIPS'25}
& \ms{78.44}{0.46}
& \ms{53.38}{1.76}
& \ms{73.03}{1.20}
& \ms{\underline{91.54}}{0.32}
& \ms{85.69}{0.77}
& \ms{69.59}{1.39}
& \ms{87.35}{0.40}
& 77.00 \\

ARCzero \venue{TKDE'26}
& \ms{79.06}{0.30}
& \ms{62.59}{1.99}
& \ms{70.46}{0.26}
& \ms{90.01}{0.20}
& \ms{\underline{87.55}}{0.26}
& \ms{67.63}{0.41}
& \ms{88.57}{0.15}
& 77.98 \\

GAD-MoRE \venue{ICDM'26}
& \ms{\underline{81.17}}{0.05}
& \ms{\underline{76.90}}{1.80}
& \ms{73.09}{0.06}
& \ms{90.28}{0.13}
& \ms{86.39}{0.20}
& \ms{\underline{75.75}}{0.95}
& \ms{\underline{91.03}}{0.11}
& \underline{82.09} \\

\midrule
\multicolumn{9}{@{}l}{\emph{Training-free Methods}} \\

TFGAD \venue{WWW'25}
& \ms{76.04}{0.19}
& \ms{70.54}{0.00}
& \ms{\textbf{74.07}}{0.05}
& \ms{74.73}{0.16}
& \ms{80.43}{0.28}
& \ms{36.49}{0.05}
& \ms{77.54}{0.00}
& 69.98 \\

FreeGAD \venue{CIKM'25}
& \ms{37.47}{0.00}
& \ms{39.44}{0.00}
& \ms{25.47}{0.00}
& \ms{71.85}{0.00}
& \ms{83.68}{0.00}
& \ms{60.41}{0.00}
& \ms{3.81}{0.00}
& 46.02 \\

\textbf{Jev-GAD (ours)}
& \ms{\textbf{95.25}}{0.10}
& \ms{\textbf{86.90}}{0.06}
& \ms{\underline{73.36}}{0.09}
& \ms{\textbf{98.41}}{0.04}
& \ms{\textbf{96.46}}{0.06}
& \ms{\textbf{80.21}}{0.11}
& \ms{\textbf{93.66}}{0.06}
& \textbf{89.18} \\

\bottomrule
\end{tabular*}
\endgroup
\end{table*}
\begin{table*}[t]
\centering
\caption{AUPRC comparison on seven graph anomaly detection benchmarks (\%). Results are reported as mean $\pm$ standard deviation. Best and second-best mean values are highlighted in \textbf{bold} and \underline{underline}, respectively.}
\label{tab:overall_auprc}

\begingroup
\setlength{\tabcolsep}{1.5pt}
\renewcommand{\arraystretch}{1.13}
\normalsize

\newcommand{\ms}[2]{#1{\footnotesize$\pm$#2}}
\newcommand{\venue}[1]{{\footnotesize(\textit{#1})}}

\begin{tabular*}{\textwidth}{@{\extracolsep{\fill}}lcccccccc@{}}
\toprule
\textbf{Method (Venue)}
& \textbf{ACM}
& \textbf{Amazon}
& \textbf{BlogCatalog}
& \textbf{CiteSeer}
& \textbf{Cora}
& \textbf{Facebook}
& \textbf{Weibo}
& \textbf{Avg.} \\
\midrule

\multicolumn{9}{@{}l}{\emph{Supervised Methods}} \\

GCN \venue{ICLR'17}
& \ms{3.68}{0.16}
& \ms{5.54}{0.51}
& \ms{5.71}{0.38}
& \ms{5.21}{0.89}
& \ms{5.85}{0.55}
& \ms{4.55}{3.06}
& \ms{25.97}{15.84}
& 8.07 \\

GAT \venue{ICLR'18}
& \ms{3.38}{0.45}
& \ms{8.10}{1.24}
& \ms{6.64}{0.71}
& \ms{4.35}{0.82}
& \ms{4.93}{0.75}
& \ms{6.18}{5.83}
& \ms{15.86}{5.83}
& 7.06 \\

BWGNN \venue{ICML'22}
& \ms{19.48}{2.20}
& \ms{6.79}{0.03}
& \ms{28.37}{4.31}
& \ms{4.02}{0.43}
& \ms{5.30}{0.87}
& \ms{5.71}{1.60}
& \ms{10.47}{0.05}
& 11.45 \\

GHRN \venue{WWW'23}
& \ms{9.21}{2.94}
& \ms{11.18}{1.10}
& \ms{24.57}{3.60}
& \ms{5.37}{0.14}
& \ms{5.23}{0.19}
& \ms{3.61}{0.68}
& \ms{8.35}{2.44}
& 9.65 \\

\midrule
\multicolumn{9}{@{}l}{\emph{Unsupervised Methods}} \\

AnomalyDAE \venue{ICASSP'20}
& \ms{11.67}{0.38}
& \ms{9.26}{1.51}
& \ms{7.17}{0.04}
& \ms{31.37}{1.09}
& \ms{32.65}{1.46}
& \ms{1.32}{0.01}
& \ms{6.10}{0.08}
& 14.22 \\

CoLA \venue{TNNLS'21}
& \ms{3.55}{0.12}
& \ms{6.55}{0.44}
& \ms{6.11}{0.36}
& \ms{4.76}{0.21}
& \ms{5.86}{0.34}
& \ms{3.41}{1.38}
& \ms{9.67}{0.50}
& 5.70 \\

TAM \venue{NeurIPS'23}
& \ms{3.74}{0.01}
& \ms{7.90}{0.05}
& \ms{5.66}{0.17}
& \ms{4.15}{0.02}
& \ms{5.58}{0.08}
& \ms{5.37}{0.51}
& \ms{6.20}{0.02}
& 5.51 \\

\midrule
\multicolumn{9}{@{}l}{\emph{Generalist Methods}} \\

ARC (10-shot) \venue{NeurIPS'24}
& \ms{36.53}{0.24}
& \ms{7.98}{1.18}
& \ms{\underline{37.77}}{0.35}
& \ms{30.93}{4.24}
& \ms{39.23}{4.40}
& \ms{4.92}{1.85}
& \ms{59.47}{2.38}
& 30.98 \\

AnomalyGFM \venue{KDD'25}
& \ms{3.10}{0.17}
& \ms{9.18}{1.69}
& \ms{4.85}{0.40}
& \ms{3.69}{0.20}
& \ms{4.36}{0.09}
& \ms{6.00}{0.52}
& \ms{9.05}{1.03}
& 5.75 \\

UNPrompt \venue{IJCAI'25}
& \ms{24.02}{1.38}
& \ms{12.28}{1.63}
& \ms{35.29}{0.90}
& \ms{9.47}{0.91}
& \ms{10.26}{0.52}
& \ms{3.44}{0.31}
& \ms{28.41}{2.25}
& 17.59 \\

IA-GGAD \venue{NeurIPS'25}
& \ms{36.33}{0.16}
& \ms{6.90}{0.29}
& \ms{36.64}{0.66}
& \ms{\underline{46.96}}{1.16}
& \ms{45.03}{2.06}
& \ms{8.65}{1.74}
& \ms{60.16}{0.93}
& 34.38 \\

ARCzero \venue{TKDE'26}
& \ms{\underline{40.62}}{0.12}
& \ms{8.53}{0.49}
& \ms{\textbf{38.03}}{0.21}
& \ms{45.03}{0.41}
& \ms{\underline{48.25}}{0.51}
& \ms{\underline{10.74}}{1.60}
& \ms{\underline{61.81}}{0.41}
& 36.14 \\

GAD-MoRE \venue{ICDM'26}
& \ms{37.17}{0.07}
& \ms{\underline{31.55}}{4.84}
& \ms{35.31}{0.42}
& \ms{40.15}{1.31}
& \ms{45.25}{0.88}
& \ms{9.00}{1.12}
& \ms{60.26}{0.30}
& \underline{36.96} \\

\midrule
\multicolumn{9}{@{}l}{\emph{Training-free Methods}} \\

TFGAD \venue{WWW'25}
& \ms{24.51}{0.73}
& \ms{19.43}{0.00}
& \ms{32.58}{0.57}
& \ms{24.34}{0.05}
& \ms{30.67}{0.63}
& \ms{1.76}{0.00}
& \ms{\textbf{65.32}}{0.00}
& 28.37 \\

FreeGAD \venue{CIKM'25}
& \ms{3.04}{0.00}
& \ms{5.14}{0.00}
& \ms{3.87}{0.00}
& \ms{19.30}{0.00}
& \ms{26.39}{0.00}
& \ms{2.98}{0.00}
& \ms{5.50}{0.00}
& 9.46 \\

\textbf{Jev-GAD (ours)}
& \ms{\textbf{61.68}}{0.29}
& \ms{\textbf{41.26}}{0.30}
& \ms{33.50}{0.25}
& \ms{\textbf{88.02}}{0.19}
& \ms{\textbf{76.10}}{0.41}
& \ms{\textbf{11.08}}{0.21}
& \ms{55.15}{0.34}
& \textbf{52.40} \\

\bottomrule
\end{tabular*}
\endgroup
\end{table*}

\subsection{Statistical Scoring and Fusion}
\label{sec:statistical_fusion}

In parallel with pretrained decision readout, Jev-GAD employs complementary statistical scorers to capture different forms of numerical irregularity in graph evidence.

\noindent\textbf{Statistical Evidence Assessment.} We apply four statistical scoring mechanisms to the heterogeneous evidence representation.

A fixed aggregation rule captures the overall magnitude of anomaly evidence, while an ECOD-style upper-tail criterion~\cite{li2022ecod} emphasizes extreme deviations in individual evidence dimensions. Isolation Forest~\cite{liu2008iforest} evaluates multivariate isolation tendencies, and an RBF-based One-Class SVM~\cite{scholkopf2001ocsvm} characterizes deviations from the estimated distribution boundary.

These scoring mechanisms operate on the same evidence representation but emphasize different forms of numerical irregularity. The statistical estimators are fitted using unlabeled target-graph evidence, without anomaly-specific supervision.

\noindent\textbf{Rank-Based Score Fusion.} Since the decision and statistical scores arise from different mechanisms, their raw values are not directly comparable. We therefore normalize each score by its empirical percentile rank within the target graph before aggregation.

Let $\mathcal{M}$ denote the set of statistical scoring mechanisms. The final anomaly score is
\begin{equation}
    s(v)=\frac{1}{|\mathcal{M}|+1}\left[\mathcal{R}_G\bigl(s_{\mathrm{dec}}(v)\bigr)+\sum_{m\in\mathcal{M}}\mathcal{R}_G\bigl(s_m(v)\bigr)\right],
    \label{eq:final_fusion}
\end{equation}
where $s_m(v)$ is the anomaly score produced by statistical mechanism $m$.

With four statistical scorers, the decision score and each statistical score contribute equally with weight $1/5$. This fusion combines pretrained decisions with complementary numerical assessments without label-dependent weight estimation.

\section{Experiments}
\label{sec:experiments}

We evaluate Jev-GAD\footnote{\url{https://github.com/zhaoxinyu2002/Jev-GAD}} on diverse attributed graphs and investigate the following research questions:

\begin{itemize}[leftmargin=*]
    \item \textbf{\textit{RQ1:}} How effective is Jev-GAD across graph domains and anomaly patterns? ($\rhd$~Section~\ref{sec:overall_performance})
    \item \textbf{\textit{RQ2:}} Do pretrained decision readout and statistical scoring provide complementary signals? ($\rhd$~Section~\ref{sec:component_ablation})
    \item \textbf{\textit{RQ3:}} Does semantic grounding improve the interpretation of numerical graph evidence? ($\rhd$~Section~\ref{sec:semantic_analysis})
    \item \textbf{\textit{RQ4:}} How do evidence views and statistical backends affect performance? ($\rhd$~Section~\ref{sec:evidence_analysis})
    \item \textbf{\textit{RQ5:}} How robust is pretrained decision readout to prompt variations and to its fusion weight? ($\rhd$~Section~\ref{sec:prompt_robustness})
    \item \textbf{\textit{RQ6:}} What are the inference-time and memory costs of Jev-GAD? ($\rhd$~Section~\ref{sec:efficiency})
\end{itemize}

\subsection{Experimental Settings}
\label{sec:exp_setup}

\noindent\textbf{Datasets and Metrics.}
We evaluate Jev-GAD on seven attributed graph anomaly detection benchmarks, following established evaluation protocols~\cite{liu2024arc,zhao2026zero,liu2022bond}: three citation graphs (\texttt{ACM}~\cite{tang2008arnetminer}, \texttt{CiteSeer} and \texttt{Cora}~\cite{sen2008collective}), three social networks (\texttt{BlogCatalog}~\cite{tang2009relational}, \texttt{Facebook} and \texttt{Weibo}~\cite{liu2024arc}), and one co-review network (\texttt{Amazon}~\cite{mcauley2013amateurs,kumar2018rev2}).
These benchmarks span diverse structures, feature distributions, and anomaly patterns.
We report AUROC~\cite{fawcett2006introduction}, AUPRC~\cite{saito2015precision}, and their seven-dataset macro-averages.

\noindent\textbf{Baselines.}
We compare against \textbf{15} methods in four categories: \emph{Supervised GAD} (GCN~\cite{kipf2017gcn}, GAT~\cite{velickovic2018gat}, BWGNN~\cite{tang2022bwgnn}, GHRN~\cite{gao2023addressing}); \emph{Unsupervised GAD} (AnomalyDAE~\cite{fan2020anomalydae}, CoLA~\cite{liu2021anomaly}, TAM~\cite{qiao2023tam}); \emph{Generalist GAD} (ARC (10-shot)~\cite{liu2024arc}, UNPrompt~\cite{niu2024zero}, AnomalyGFM~\cite{qiao2025anomalygfm}, IA-GGAD~\cite{zhang2026ia}, ARCzero~\cite{liu2026arczero}, GAD-MoRE~\cite{zhao2026zero}); and \emph{Training-free GAD} (TFGAD~\cite{zhou2025training}, FreeGAD~\cite{zhao2025freegad}).
Source-trained comparisons use disjoint source and target graphs following established cross-graph protocols~\cite{zhao2026zero}, with source datasets detailed in Appendix~\ref{app:datasets}.
Previously reported and reproduced results follow their respective protocols. In particular, ARC (10-shot) uses target labels and serves as a label-assisted reference rather than a strictly label-free baseline.

\noindent\textbf{Evaluation Protocol and Implementation.}
Jev-GAD requires neither target anomaly labels nor GAD-specific gradient-based optimization at inference time.
Anomaly labels are used only for metric computation, while statistical estimators operate on unlabeled target-graph evidence.
We use \textbf{Qwen3-4B}~\cite{yang2025qwen3} as the frozen backbone for non-generative decision readout, extracting scores from two fixed candidate decision tokens without autoregressive generation or parameter updates. 

The evidence representation combines structural, attribute, and neighborhood statistics from the raw and aligned feature views, with the aligned dimension set to 32.
The decision state retains structural rarity, aligned-view attribute deviation, and aligned-view neighborhood inconsistency.
The statistical branch consists of fixed aggregation, ECOD-style upper-tail scoring~\cite{li2022ecod}, Isolation Forest (IF)~\cite{liu2008iforest}, and RBF-based One-Class SVM (OCSVM)~\cite{scholkopf2001ocsvm}.
IF uses 1,000 trees, and its stochastic results are averaged over five random seeds.
All five scores undergo graph-relative rank normalization before equal-weight fusion.

\begin{figure*}[!t]
    \centering
    \includegraphics[width=\textwidth]{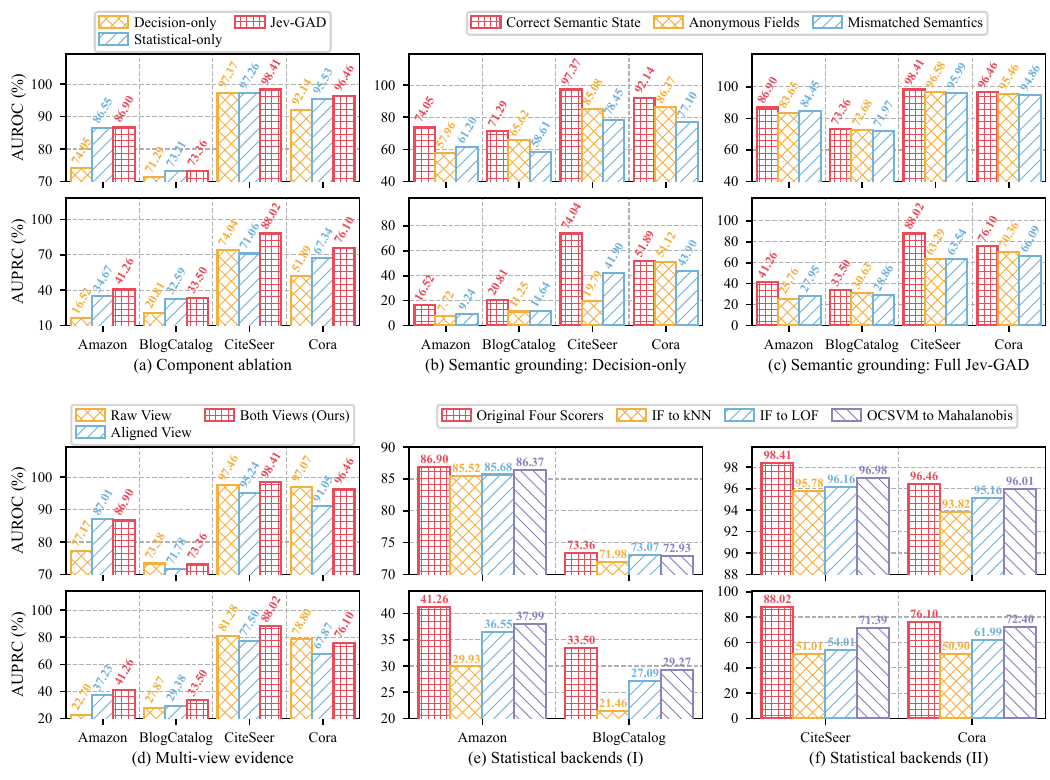}
    \Description{Six grouped bar charts of AUROC and AUPRC on Amazon, BlogCatalog, CiteSeer, and Cora: component ablation; semantic grounding under decision-only and full Jev-GAD scoring; raw, aligned, and combined evidence views; and one-for-one statistical backend replacements on two pairs of graphs.}
    \caption{Diagnostic analyses on four graphs. (a) Component ablation. (b, c) Semantic grounding under Decision-only and full Jev-GAD scoring; Mismatched Semantics averages three fixed cyclic shifts. (d) Raw, aligned, and combined evidence views. (e, f) One-for-one replacement of statistical scorers. Y-axes do not all start at zero; numbers above bars are exact scores (\%).}
    \label{fig:analysis}
\end{figure*}

\subsection{\textit{RQ1:} Overall Effectiveness}
\label{sec:overall_performance}
Tables~\ref{tab:overall_auroc} and~\ref{tab:overall_auprc} compare Jev-GAD against the baselines in terms of AUROC and AUPRC.
The results reveal the following findings.

\noindent\textbf{(1) Strong overall performance.}
Jev-GAD achieves average AUROC and AUPRC scores of \textbf{89.18\%} and \textbf{52.40\%}, outperforming the strongest baseline macro-averages by 7.09 and 15.44 percentage points, respectively.
It obtains the highest AUROC on six of seven graphs and the highest AUPRC on five, demonstrating consistently strong performance across diverse graph benchmarks.

\noindent\textbf{(2) Pronounced gains on selected graph domains.}
On \texttt{CiteSeer} and \texttt{Cora}, Jev-GAD obtains AUPRC scores of 88.02\% and 76.10\%, respectively.
These improvements demonstrate the effectiveness of integrating heterogeneous anomaly evidence with pretrained decision readout to capture diverse anomaly characteristics.

\subsection{\textit{RQ2:} Component Ablation}
\label{sec:component_ablation}
We compare three configurations: \textbf{(1) Decision-only}, using the pretrained score; \textbf{(2) Statistical-only}, combining four numerical scorers; and \textbf{(3) Jev-GAD}, fusing both pathways.
Figure~\ref{fig:analysis}(a) reports their AUROC and AUPRC on four diagnostic datasets.

\noindent\textbf{(1) Decision readout and numerical scoring are complementary.}
Jev-GAD outperforms both pathways on all four datasets.
On \texttt{CiteSeer}, Decision-only, Statistical-only, and Jev-GAD achieve AUPRC scores of 74.04\%, 71.06\%, and 88.02\%; on \texttt{Cora}, the scores are 51.89\%, 67.34\%, and 76.10\%.
These gains confirm complementary anomaly-ranking information from both pathways.

\noindent\textbf{(2) The value of pretrained decisions varies by dataset.}
Compared with Statistical-only, Jev-GAD improves AUPRC by 6.59, 0.91, 16.96, and 8.76 percentage points on \texttt{Amazon}, \texttt{BlogCatalog}, \texttt{CiteSeer}, and \texttt{Cora}, respectively. These varying gains reflect the dataset-dependent contribution of pretrained decisions across heterogeneous anomaly patterns.

\noindent\textbf{(3) Pretrained decisions provide unique value beyond additional numerical scorers.}
To distinguish semantic decision-making from the effect of introducing an additional scoring signal, we retain the four statistical scorers and compare the pretrained decision score against three numerical alternatives derived from the same seven-dimensional evidence: Mean-7D, Max-7D, and LOF-7D.
Each five-score configuration uses equal-weight rank fusion.
Across the four diagnostic graphs, Jev-GAD achieves a macro-AUPRC of \textbf{59.72\%}, outperforming Statistical-only (51.41\%), Mean-7D (49.98\%), Max-7D (50.04\%), and LOF-7D (42.14\%).
The corresponding macro-AUROC reaches 88.78\%, compared with 88.14\% for Statistical-only.
These results demonstrate that pretrained semantic decisions contribute complementary anomaly-ranking information beyond that provided by additional numerical scoring mechanisms.
Detailed comparisons are reported in Table~\ref{tab:app:alternative_fifth}.

\begin{figure}[t]
    \centering
    \includegraphics[width=\columnwidth]{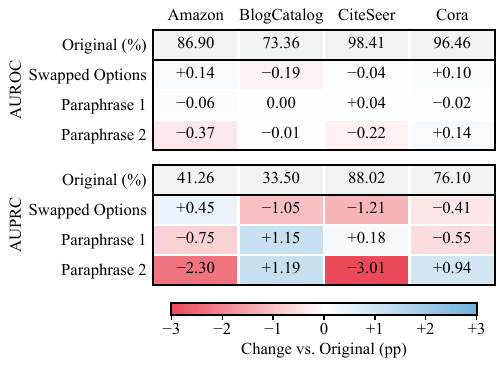}
    \caption{Prompt robustness. The first row of each block gives the original score (\%); the other rows give changes relative to it in percentage points.}
    \label{fig:prompt_robustness}
\end{figure}

\begin{figure}[t]
    \centering
    \includegraphics[width=\columnwidth]{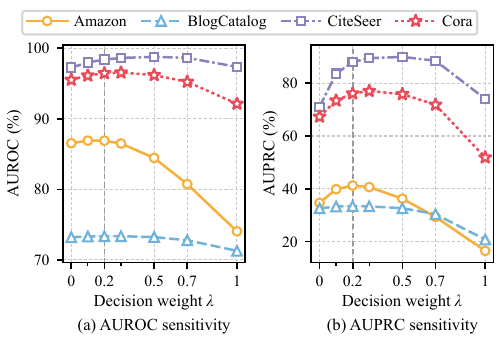}
    \caption{Sensitivity to decision weight $\lambda$ on four graphs. The dashed line marks equal-weight fusion ($\lambda=0.2$), while shaded bands indicate $\pm1$ standard deviation across five Isolation Forest seeds. The endpoints represent Statistical-only ($\lambda=0$) and Decision-only ($\lambda=1$).}
    \label{fig:fusion_weight}
\end{figure}

\begin{figure}[t]
    \centering
    \includegraphics[width=\columnwidth]{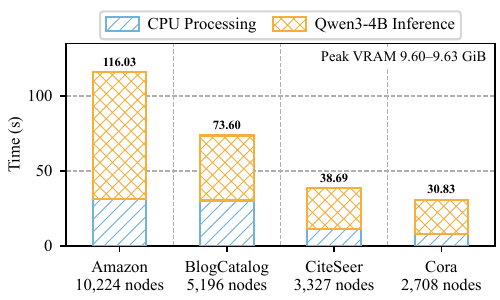}
    \Description{Stacked bars of CPU processing time and frozen Qwen3-4B inference time on four graphs, with measured totals above the bars and node counts below.}
    \caption{Computational efficiency. Numbers above bars are measured total times (s).}
    \label{fig:efficiency}
\end{figure}

\subsection{\textit{RQ3:} Semantic Evidence Grounding}
\label{sec:semantic_analysis}
We test whether pretrained decisions depend on the correspondence between numerical evidence and its semantic description. Two controlled variants are considered: \textbf{(1) Anonymous Fields}, which replace evidence descriptors with anonymous identifiers while preserving all values; and \textbf{(2) Mismatched Semantics}, which cyclically reassign values to incorrect descriptors. The latter is averaged over three predefined permutations. Figure~\ref{fig:analysis}(b) and (c) compare the pretrained decision pathway and the final fused scores, respectively.

\noindent\textbf{(1) Semantic correspondence matters for decision readout.}
On \texttt{CiteSeer}, Decision-only AUPRC declines from 74.04\% with the original descriptions to 19.79\% with Anonymous Fields and 41.90\% with Mismatched Semantics. Similar declines across other datasets highlight the importance of semantic descriptions and their correct alignment with numerical evidence.

\noindent\textbf{(2) The benefit survives statistical fusion.}
On \texttt{Amazon}, the complete Jev-GAD model reaches 41.26\% AUPRC with the original state, compared with 25.76\% and 27.95\% under Anonymous Fields and Mismatched Semantics. The observed trends across the four diagnostic graphs support the role of explicit semantic grounding beyond its effect on the standalone decision branch. Because the numerical evidence and four statistical scores remain unchanged under both perturbations, the performance degradation isolates the contribution of semantic grounding to the fused ranking.

\subsection{\textit{RQ4:} Evidence and Scoring Analysis}
\label{sec:evidence_analysis}
We assess two design choices that affect how heterogeneous numerical evidence is used: feature-view composition and the statistical scoring backend.

\noindent\textbf{(1) Raw and aligned views contain complementary evidence.}
Figure~\ref{fig:analysis}(d) compares \emph{Raw View}, \emph{Aligned View}, and \emph{Both Views} while keeping pretrained decision readout unchanged. Combining both views produces the highest AUPRC on \texttt{Amazon}, \texttt{BlogCatalog}, and \texttt{CiteSeer}. On \texttt{Amazon}, the three configurations obtain 22.70\%, 37.23\%, and 41.26\% AUPRC, respectively. Combining both views achieves the highest AUPRC on three of the four datasets, demonstrating the effectiveness of complementary anomaly evidence.

\noindent\textbf{(2) Performance depends on the statistical backend.}
We compare the original ensemble with three alternative configurations: replacing IF with kNN or LOF, and replacing OCSVM with Mahalanobis distance. As shown in Figure~\ref{fig:analysis}(e) and (f), the original backend yields higher AUROC and AUPRC than these variants on all four diagnostic datasets. For example, its AUPRC on \texttt{CiteSeer} is 88.02\%, compared with 51.01\%, 54.01\%, and 71.39\% for the tested alternatives. These results demonstrate the effectiveness of the proposed statistical scoring configuration.

\subsection{\textit{RQ5:} Decision Robustness}
\label{sec:prompt_robustness}
We examine prompt sensitivity by swapping option order or using two alternative decision instructions while keeping the evidence states fixed.

\noindent\textbf{(1) Robustness to prompt variations.}
Figure~\ref{fig:prompt_robustness} shows that Jev-GAD achieves 88.02\% AUPRC on \texttt{CiteSeer} with the original prompt and 86.81\%, 88.20\%, and 85.01\% under three variants. On \texttt{Cora}, AUPRC ranges from 75.55\% to 77.04\%. These results demonstrate stable performance across the tested prompt formulations and option orderings.

\noindent\textbf{(2) Sensitivity to decision weight.}
We vary the decision weight $\lambda\in\{0,\allowbreak 0.1,\allowbreak 0.2,\allowbreak 0.3,\allowbreak 0.5,\allowbreak 0.7,\allowbreak 1\}$, distributing the remaining weight equally among four statistical scorers. Here, $\lambda=0$ and $\lambda=1$ correspond to Statistical-only and Decision-only, while $\lambda=0.2$ recovers equal-weight fusion. Figure~\ref{fig:fusion_weight} shows peak AUPRC at $\lambda=0.2$ on \texttt{Amazon} and \texttt{BlogCatalog}, 0.3 on \texttt{Cora}, and 0.5 on \texttt{CiteSeer}. Increasing $\lambda$ from 0.2 to 0.3 changes macro-AUPRC by only 0.41 points (59.72\% to 60.13\%), whereas the endpoints lose 8.31 and 18.90 points relative to the default. The default also achieves the highest macro-AUROC.

\subsection{\textit{RQ6:} Computational Efficiency}
\label{sec:efficiency}
We measure graph evidence processing and frozen-model decision inference separately, as summarized in Figure~\ref{fig:efficiency}.

\noindent\textbf{(1) Runtime includes preprocessing and model inference.}
On \texttt{Amazon} (10,224 nodes), Jev-GAD takes approximately 116 seconds, including 31 seconds of CPU processing and 85 seconds of pretrained decision inference. On \texttt{Cora} (2,708 nodes), it takes approximately 31 seconds.

\noindent\textbf{(2) Pretrained decisions introduce a measurable cost.}
Frozen Qwen3-4B readout processes approximately 120 nodes per second with about 9.6 GiB of peak allocated GPU memory. This is the primary trade-off relative to numerical training-free scoring, which requires no language-model inference.

\section{Conclusion}
\label{sec:conclusion}

We presented Jev-GAD, a training-free generalist graph anomaly detection framework connecting heterogeneous anomaly evidence with a single frozen pretrained model. The semantic Graph-to-State (G2S) transformation organizes graph-relative evidence into semantically grounded inputs for non-generative decision readout, while rank-based fusion integrates pretrained decisions with complementary statistical scores. Across seven benchmarks, Jev-GAD achieves average AUROC and AUPRC of 89.18\% and 52.40\%, respectively. Component ablations and robustness analyses validate the effectiveness of pretrained decision readout and semantic grounding for training-free graph anomaly detection.


\bibliographystyle{ACM-Reference-Format}
\bibliography{sample-base}

\appendix
\nobalance
\appendix

\section{Additional Method Details}
\label{app:method}

\subsection{Graph Evidence Construction}
\label{app:evidence}
Jev-GAD first converts graph statistics into directionally consistent, graph-relative evidence.
We remove self-loops and symmetrize the adjacency matrix, treating an edge in either direction as an undirected connection.
For a statistic $q_j(v)$, we use the empirical mid-rank percentile
\begin{equation}
  R_G(q_j(v))=\frac{\operatorname{rank}_{\mathrm{avg}}(q_j(v))-1}{|V|-1},
  \label{eq:app:rank}
\end{equation}
where tied values receive their average rank.
For structural quantities whose atypicality can occur at either tail, we rank $|q_j(v)-\operatorname{median}_{u\in V}q_j(u)|$ instead.

Table~\ref{tab:app:state_fields} summarizes the seven evidence entries passed to the pretrained decision model.
All features are measured on the input graph, and larger percentiles represent stronger deviations on the corresponding statistic.

\begin{table}[H]
\centering
\caption{Node-level evidence fields in the semantic decision state.}
\label{tab:app:state_fields}
\footnotesize

\setlength{\fboxsep}{0pt}

\begin{tcolorbox}[
    width=0.96\linewidth,
    colback=white,
    colframe=black,
    boxrule=0.6pt,
    arc=1mm,
    left=2mm,
    right=2mm,
    top=1.5mm,
    bottom=1.5mm
]

\begin{tcolorbox}[
    colback=gray!4,
    colframe=black,
    boxrule=0.4pt,
    arc=0.8mm,
    left=1.5mm,
    right=1.5mm,
    top=1mm,
    bottom=1mm,
    enhanced,
    before skip=0pt,
    after skip=2mm
]
\textbf{Neighborhood} \hfill \emph{aligned feature view}

\vspace{1mm}
\begin{itemize}[leftmargin=1.6em, itemsep=1pt, topsep=0pt, parsep=0pt]
  \item \texttt{neighbor\_dissimilarity\_percentile}
  \item \texttt{neighbor\_diversity\_percentile}
\end{itemize}
\end{tcolorbox}

\begin{tcolorbox}[
    colback=gray!4,
    colframe=black,
    boxrule=0.4pt,
    arc=0.8mm,
    left=1.5mm,
    right=1.5mm,
    top=1mm,
    bottom=1mm,
    enhanced,
    before skip=0pt,
    after skip=2mm
]
\textbf{Structural} \hfill \emph{original topology}

\vspace{1mm}
\begin{itemize}[leftmargin=1.6em, itemsep=1pt, topsep=0pt, parsep=0pt]
  \item \texttt{degree\_rarity}
  \item \texttt{clustering\_rarity}
  \item \texttt{core\_number\_rarity}
  \item \texttt{pagerank\_rarity}
\end{itemize}
\end{tcolorbox}

\begin{tcolorbox}[
    colback=gray!4,
    colframe=black,
    boxrule=0.4pt,
    arc=0.8mm,
    left=1.5mm,
    right=1.5mm,
    top=1mm,
    bottom=1mm,
    enhanced,
    before skip=0pt,
    after skip=0pt
]
\textbf{Attribute} \hfill \emph{aligned feature view}

\vspace{1mm}
\begin{itemize}[leftmargin=1.6em, itemsep=1pt, topsep=0pt, parsep=0pt]
  \item \texttt{centroid\_l2\_distance\_percentile}
\end{itemize}
\end{tcolorbox}

\end{tcolorbox}
\end{table}

The decision state uses the aligned feature view, while the numerical scorers receive a more extensive evidence matrix.
Specifically, the scoring matrix contains four structural columns, six attribute and neighborhood columns from the original features, and six from the aligned features, yielding $4+6+6=16$ evidence columns.
The extra attribute measurements include feature magnitude or cosine deviation, whitened distance, maximum standardized deviation, and residual energy, as applicable to the feature view.
All columns are converted into graph-relative percentiles before statistical scoring.

\paragraph{Feature alignment.} The 32-dimensional aligned representation contains a 16-dimensional Euclidean view and four 4-dimensional curvature views, with curvatures $\{-1,+1,-0.5,+0.5\}$.
Each curvature view is constructed using manifold projection, exponential mapping, logarithmic mapping, and candidate selection, following the alignment construction in~\cite{zhao2026zero}.
Candidate features are obtained by PCA when necessary and ranked by normalized graph Laplacian scores~\cite{he2005laplacian}; the lowest scores are retained for each view.
This preprocessing uses a fixed random seed and does not optimize anomaly-specific model parameters.

\subsection{Semantic Graph-to-State Interface}
\label{app:g2s}
Each node $v$ is represented as a pair $S_v=(c_G,B_v)$, corresponding to the \texttt{graph\_context} and \texttt{node\_evidence} objects.
The graph-level context $c_G$ contains the node and edge counts, edge density, mean degree, aligned feature dimension, and mean node--neighbor cosine similarity.
The seven node-level percentiles in $B_v$ are grouped in the fixed order \emph{neighborhood}, \emph{structural}, and \emph{attribute}.
Numeric values are serialized to six decimal places.
This fixed organization preserves the correspondence between an observation and its anomaly-related meaning.

\paragraph{Decision query and readout.} The serialized state forms the evidence part of a fixed structured query.
The criterion instructs the model to interpret all node-level values as graph-relative anomaly percentiles.
Two labeled alternatives are provided: \texttt{A}: anomalous behavior, and \texttt{B}: ordinary node variation.
The query presents the criterion, the candidate options, and then the node-specific evidence.
The frozen \texttt{Qwen/Qwen3-4B} model processes the query with thinking disabled and without autoregressive answer generation.
We extract the logits $\ell_A(v)$ and $\ell_B(v)$ at the answer position, yielding
\begin{equation}
 s_{\mathrm{dec}}(v)
 =\frac{\exp(\ell_A(v))}{\exp(\ell_A(v))+\exp(\ell_B(v))}.
 \label{eq:app:readout}
\end{equation}
The backbone is loaded in FP16; option logits are computed in FP32.
A common query prefix is reused across nodes to reduce repeated forward computation.
Neither the backbone nor the decision interface is adapted with GAD-specific gradients.

\subsection{Statistical Scoring and Rank Fusion}
\label{app:backends}
The four numerical scoring rules operate on the 16-dimensional percentile evidence matrix $E\in[0,1]^{N\times16}$: \emph{Fixed} averages the 16 evidence values; \emph{ECOD}~\cite{li2022ecod} aggregates their upper-tail evidence, $-\sum_j\log(1-E_{vj}+0.5/N)$; \emph{Isolation Forest}~\cite{liu2008iforest} uses 1,000 estimators; and \emph{OCSVM}~\cite{scholkopf2001ocsvm} uses an RBF kernel with \texttt{gamma=scale}.
Isolation Forest uses five seeds $\{0,1,2,3,4\}$; the other scorers are deterministic under the evaluation protocol.
Every score is individually converted into its empirical percentile rank before fusion.
The final score is the arithmetic mean of the decision rank and the four numerical ranks, as defined in Eq.~\eqref{eq:final_fusion}.
Thus the default decision weight is $1/5$, as examined in the fusion sensitivity experiment in the main text (Figure~\ref{fig:fusion_weight}).

\renewcommand{\thealgorithm}{A\arabic{algorithm}}
\setcounter{algorithm}{0}

\begin{algorithm}[t]
\caption{Training-Free Scoring with Jev-GAD}
\label{alg:jev_gad}
\footnotesize

\begin{algorithmic}[1]
\Require Attributed graph $G=(V,E,X_0)$; frozen decision model $g_\theta$
\Ensure Node-level anomaly scores $\{s(v)\}_{v\in V}$

\Statex \textbf{Stage I: Graph Evidence Construction}
\State $\widetilde X \gets \operatorname{Align}_{32}(G,X_0)$
\State $\mathbf E_{16} \gets \operatorname{BuildEvidence}(G,X_0,\widetilde X)$

\Statex \textbf{Stage II: Semantic Decision Readout}
\ForAll{$v\in V$}
    \State $S_v \gets (c_G,\operatorname{Select}_{7}(\mathbf E_{16}[v]))$
    \State $s_{\mathrm{dec}}(v) \gets \operatorname{Readout}_{A/B}(g_\theta,S_v)$
\EndFor

\Statex \textbf{Stage III: Statistical Scoring and Fusion}
\State $\mathcal M \gets \{\mathrm{Fixed},\mathrm{ECOD},\mathrm{IF},\mathrm{OCSVM}\}$
\ForAll{$m\in\mathcal M$}
    \State $s_m \gets \operatorname{Score}_{m}(\mathbf E_{16})$
\EndFor

\ForAll{$v\in V$}
    \State $s(v)\gets \dfrac{R_G(s_{\mathrm{dec}}(v))+\sum_{m\in\mathcal M}R_G(s_m(v))}{5}$
\EndFor

\end{algorithmic}
\end{algorithm}

\section{Experiment Details}
\label{app:settings}

\subsection{Dataset Details}
\label{app:datasets}
We evaluate Jev-GAD on seven attributed graph anomaly detection benchmarks spanning three domains, following established GAD benchmark protocols~\cite{liu2024arc,qiao2023tam,zhao2026zero}.
For source-trained baseline comparisons, the cross-graph protocol additionally uses four disjoint source graphs: \texttt{PubMed}~\cite{sen2008collective}, \texttt{Flickr}~\cite{ding2019deep,tang2009relational}, \texttt{Reddit}~\cite{kumar2019predicting}, and \texttt{YelpChi}~\cite{rayana2015collective,mcauley2013amateurs}.
These source graphs are used only by the applicable baselines and are not used to train or configure Jev-GAD.
Table~\ref{tab:app:datasets} reports statistics of the seven target graphs used for evaluation.

\begin{itemize}[leftmargin=*]
  \item \textbf{Citation Domain:} \texttt{ACM}, \texttt{CiteSeer}, and \texttt{Cora}~\cite{ding2019deep,liu2024arc}, where nodes represent papers and edges denote citation links.
  Node attributes are derived from paper content, and the benchmark versions contain injected structural and attribute anomalies.
  \item \textbf{Social Domain:} \texttt{BlogCatalog}~\cite{qiao2023tam} is a social-blogging network with user profile and content features and injected anomalies; \texttt{Facebook} and \texttt{Weibo}~\cite{liu2024arc} are social networks with annotated anomalous accounts or behaviors.
  \item \textbf{Co-review Domain:} \texttt{Amazon}~\cite{kumar2018rev2,rayana2015collective} connects reviewers through co-review activity, and its anomalies arise from suspicious review patterns.
\end{itemize}
Anomaly labels are used only for metric computation on the seven evaluation graphs.

\begin{table}[H]
\centering
\caption{Statistics of the seven benchmark graphs. Edges are undirected and self-loop-free; Ratio is the percentage of anomalous nodes and Deg.\ the average degree.}
\label{tab:app:datasets}
\footnotesize
\setlength{\tabcolsep}{2.4pt}
\begin{tabular}{@{}llrrrrrr@{}}
\toprule
Dataset & Domain & \#Node & \#Edge & \#Feat. & \#Anom. & Ratio & Deg. \\
\midrule
ACM & Citation & 16,484 & 74,073 & 8,337 & 597 & 3.62 & 8.99 \\
CiteSeer & Citation & 3,327 & 5,077 & 3,703 & 150 & 4.51 & 3.05 \\
Cora & Citation & 2,708 & 5,803 & 1,433 & 150 & 5.54 & 4.29 \\
BlogCatalog & Social & 5,196 & 172,783 & 8,189 & 300 & 5.77 & 66.51 \\
Facebook & Social & 1,081 & 27,552 & 576 & 25 & 2.31 & 50.98 \\
Weibo & Social & 8,405 & 377,271 & 400 & 868 & 10.33 & 89.77 \\
Amazon & Co-review & 10,224 & 175,608 & 25 & 693 & 6.78 & 34.35 \\
\bottomrule
\end{tabular}
\end{table}

\subsection{Baseline Details}
\label{app:baselines}
We compare Jev-GAD with 15 baselines from four categories.
\begin{itemize}[leftmargin=*]
  \item \textbf{Supervised GAD:} GCN~\cite{kipf2017gcn} and GAT~\cite{velickovic2018gat}, representative graph neural networks based on neighborhood aggregation and attention, and BWGNN~\cite{tang2022bwgnn} and GHRN~\cite{gao2023addressing}, two GAD-oriented graph learning methods.
  All four are trained with labeled examples.
  \item \textbf{Unsupervised GAD:} AnomalyDAE~\cite{fan2020anomalydae}, which reconstructs graph structure and attributes; CoLA~\cite{liu2021anomaly}, which contrasts nodes with their contextual subgraphs; and TAM~\cite{qiao2023tam}, which models node--neighbor affinity.
  \item \textbf{Generalist GAD:} UNPrompt~\cite{niu2024zero}, AnomalyGFM~\cite{qiao2025anomalygfm}, and IA-GGAD~\cite{zhang2026ia}, which learn transferable anomaly detection capabilities across graph domains; ARCzero~\cite{liu2026arczero}, evaluated in its zero-shot transfer setting; GAD-MoRE~\cite{zhao2026zero}, which models cross-graph anomaly evidence with a mixture of geometric experts; and ARC (10-shot)~\cite{liu2024arc}, which adapts across graphs with in-context examples and serves as a label-assisted reference rather than a strict label-free baseline.
  Methods with prior training on source graphs are grouped separately from training-free methods.
  \item \textbf{Training-free GAD:} TFGAD~\cite{zhou2025training} and FreeGAD~\cite{zhao2025freegad}, the two dedicated training-free baselines, which perform no anomaly-specific parameter training on the target graph.
\end{itemize}

\noindent\textbf{Source-Graph Training Protocol.}
Source-trained baselines following the cross-graph protocol of~\cite{zhao2026zero} are trained on the four source graphs described in Appendix~\ref{app:datasets} and evaluated on seven disjoint target graphs.
Supervised baselines use source-graph anomaly labels, whereas unsupervised baselines train without them.
Previously reported results are taken from their respective published evaluations where applicable.
We reproduce ARCzero, TFGAD, and FreeGAD with their available implementations on the graph versions used by Jev-GAD, keeping the implementation settings instead of tuning individual test graphs; ARCzero uses five trials, and TFGAD and FreeGAD use five random seeds.

\subsection{Implementation Details}
\label{app:implementation}
We introduce the general implementation details below.

\noindent\textbf{Decision readout.} The frozen \texttt{Qwen/}\allowbreak\texttt{Qwen3-4B} backbone is loaded in FP16 with thinking disabled.
The two option logits are computed in FP32 at the answer position without autoregressive decoding, and decision inference uses a batch size of 32.
The inference cache is keyed by the backbone, numerical precision, query template, and node-specific prompt digest, so the rank fusion and the weight-sensitivity analysis can be recomputed from saved scores without repeating language-model inference.

\noindent\textbf{Scoring and evaluation.} Graph preprocessing, feature alignment, evidence construction, and statistical scoring run independently for each graph on the CPU; feature alignment and residual-energy computation are restricted to one thread.
Isolation Forest results are averaged over five seeds, and the main tables report mean and standard deviation.
AUROC and AUPRC are computed from node-level anomaly scores; labels are not required to construct target-graph evidence, decision states, or inference scores.

\noindent\textbf{Environment.}\label{app:hardware} Experiments are conducted with:
\begin{itemize}[leftmargin=*]
  \item \textbf{Operating System:} Ubuntu 24.04.3 LTS (Linux kernel 6.8).
  \item \textbf{CPU:} two AMD EPYC 7742 64-core processors (256 threads) with 1007~GiB of memory.
  \item \textbf{GPU:} eight NVIDIA A100-SXM4-80GB GPUs (driver 580.65.06), one GPU per inference process.
  The efficiency measurements in Section~\ref{sec:efficiency} use one GPU, processing about 120 nodes per second with about 9.6~GiB of peak allocated memory.
  \item \textbf{Software:} CUDA 13.0, Python 3.12.3, PyTorch 2.14.1, Transformers 5.19.0, NumPy 2.5.3, SciPy 1.18.1, pandas 3.0.6, scikit-learn 1.9.1, NetworkX 3.7, and Geoopt 0.5.1.
\end{itemize}

\section{Additional Experimental Analysis}
\label{app:additional}

\subsection{Decision-State Evidence Granularity}
\label{app:granularity}
Following the analysis of semantic grounding in the main text, we further investigate the number of evidence entries supplied to the decision model.
The \emph{5D} state retains two neighborhood cues, two structural rarity cues (degree and PageRank), and one aligned attribute-distance cue.
The default \emph{7D} state adds clustering and core-number rarity.
The \emph{10D} state additionally introduces whitened distance, maximum standardized deviation, and residual energy.
All variants retain the same scoring branch and fusion protocol.

\begin{table}[H]
\centering
\caption{AUPRC (\%) of the fused method with different semantic state sizes. The 7D configuration is the default Jev-GAD setting.}
\label{tab:app:granularity}
\footnotesize
\setlength{\tabcolsep}{5pt}
\begin{tabular}{@{}lrrr@{}}
\toprule
Dataset & 5D & 7D & 10D \\
\midrule
Amazon & 37.48 & 41.26 & \textbf{42.34} \\
BlogCatalog & 32.86 & 33.50 & \textbf{33.73} \\
CiteSeer & 85.29 & \textbf{88.02} & 86.23 \\
Cora & 74.12 & \textbf{76.10} & 73.58 \\
\midrule
Macro average & 57.44 & \textbf{59.72} & 58.97 \\
\bottomrule
\end{tabular}
\end{table}

As shown in Table~\ref{tab:app:granularity}, the seven-field representation achieves the highest macro-AUPRC across the four diagnostic datasets.
Expanding the state from 5D to 7D improves all four datasets.
Extending it to 10D provides small additional gains on Amazon and BlogCatalog but reduces performance on CiteSeer and Cora.
The analysis illustrates that decision quality depends on organizing the most useful semantic evidence, rather than simply increasing the number of numerical fields passed to the pretrained model.

\subsection{Cumulative Anomaly Retrieval}
\label{app:case}

\begin{figure}[H]
    \centering
    \includegraphics[width=0.85\columnwidth]{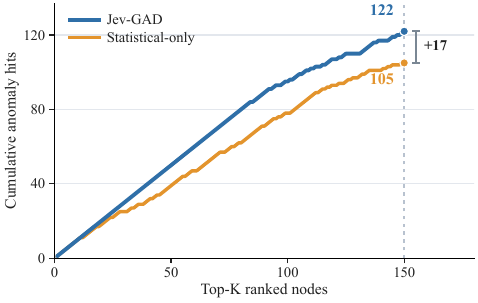}
    \Description{Cumulative anomaly hits across ranking depths on CiteSeer, comparing Jev-GAD with Statistical-only. At Top-150, Jev-GAD retrieves 122 anomalies, compared with 105 for Statistical-only.}
    \caption{Cumulative anomaly hits at different ranking depths on CiteSeer. Jev-GAD retrieves 122 true anomalies at Top-150, compared with 105 for Statistical-only.}
    \label{fig:app:case_study}
\end{figure}

We examine how pretrained decision scores improve anomaly prioritization by comparing cumulative anomaly retrieval under Statistical-only scoring and Jev-GAD on \texttt{CiteSeer}.
As shown in Figure~\ref{fig:app:case_study}, Jev-GAD retrieves 122 true anomalies among the top-150 ranked nodes, compared with 105 for Statistical-only.
This improvement of 17 additional detections demonstrates the benefit of incorporating pretrained decisions into anomaly ranking.

\subsection{Alternative Fifth-Scorer Controls}
\label{app:alternative_fifth}

We examine whether pretrained decision scores can be replaced by additional numerical scoring mechanisms.
The original four statistical scorers remain fixed, while the fifth score is selected from Mean-7D, Max-7D, LOF-7D, or pretrained semantic decisions.
All five-score configurations employ equal-weight rank fusion, while Statistical-only uses the original four-score aggregation.

\begin{table}[H]
\centering
\caption{Alternative fifth-scorer controls on four diagnostic datasets (AUPRC, \%).}
\label{tab:app:alternative_fifth}
\footnotesize
\setlength{\tabcolsep}{2pt}
\begin{tabular}{@{}lrrrrr@{}}
\toprule
Dataset & Stat. & Mean & Max & LOF & Semantic \\
\midrule
Amazon      & 34.67 & 28.11 & 25.80 & 30.93 & \textbf{41.26} \\
BlogCatalog & 32.59 & 30.23 & \textbf{33.80} & 28.87 & 33.50 \\
CiteSeer    & 71.06 & 70.29 & 71.36 & 56.13 & \textbf{88.02} \\
Cora        & 67.34 & 71.29 & 69.20 & 52.63 & \textbf{76.10} \\
\midrule
Macro Avg.  & 51.41 & 49.98 & 50.04 & 42.14 & \textbf{59.72} \\
\bottomrule
\end{tabular}
\end{table}

As shown in Table~\ref{tab:app:alternative_fifth}, pretrained semantic decisions achieve the highest macro-AUPRC of 59.72\%, exceeding Statistical-only (51.41\%) and all three numerical fifth-scorer alternatives.
The gains are particularly pronounced on \texttt{CiteSeer} and \texttt{Cora}, while Max-7D performs slightly better on \texttt{BlogCatalog}.
Overall, pretrained decisions provide complementary anomaly-ranking information beyond additional numerical scoring.

\subsection{Conditional Ranking Beyond Statistical Scores}
\label{app:conditional_ranking}

We further examine whether pretrained decisions distinguish anomalous nodes with similar statistical anomaly scores.
For each graph, nodes are partitioned into ten equal-frequency bins according to their Statistical-only scores, with tied scores assigned to the same bin.
Within bins containing both anomalous and normal nodes, we compute pairwise ranking AUC for each scoring method and aggregate the results weighted by the number of valid anomaly--normal pairs.

\begin{table}[H]
\centering
\caption{Pair-weighted within-bin AUC conditioned on Statistical-only anomaly scores.}
\label{tab:app:conditional_ranking}
\footnotesize
\setlength{\tabcolsep}{2pt}
\begin{tabular}{@{}lrrrrr@{}}
\toprule
Dataset & Stat. & Mean & Max & LOF & Semantic \\
\midrule
Amazon      & 0.592 & 0.353 & 0.314 & 0.511 & \textbf{0.611} \\
BlogCatalog & \textbf{0.613} & 0.495 & 0.566 & 0.544 & 0.593 \\
CiteSeer    & 0.783 & 0.660 & 0.703 & 0.491 & \textbf{0.914} \\
Cora        & 0.750 & 0.747 & 0.679 & 0.512 & \textbf{0.813} \\
\midrule
Macro Avg.  & 0.684 & 0.564 & 0.565 & 0.514 & \textbf{0.733} \\
\bottomrule
\end{tabular}
\end{table}

Table~\ref{tab:app:conditional_ranking} shows that pretrained semantic decisions achieve a macro within-bin AUC of 0.733, compared with 0.684 for Statistical-only and 0.564, 0.565, and 0.514 for Mean-7D, Max-7D, and LOF-7D, respectively.
The improvements are most pronounced on \texttt{CiteSeer} and \texttt{Cora}, demonstrating that semantic decisions provide additional discriminative information even among nodes with similar statistical scores.

The analysis considers only bins containing both anomalous and normal nodes, covering 6.68\%--8.99\% of all anomaly--normal pairs across the four graphs.

\end{document}